\documentclass{article}

\usepackage[nonatbib, final]{nips_2018}

\usepackage[utf8]{inputenc} 
\usepackage[T1]{fontenc}    
\usepackage{hyperref}       
\usepackage{url}            
\usepackage{booktabs}       
\usepackage{amsfonts}       
\usepackage{nicefrac}       
\usepackage{microtype}      

\usepackage{graphicx}
\usepackage{amsmath,amssymb} 

\usepackage{multirow}
\usepackage{graphbox} 
\usepackage{booktabs} 
\usepackage{tabularx} 
\usepackage{moresize}

\newcommand{\bx}{\mathbf{x}}
\newcommand{\by}{\mathbf{y}}
\newcommand{\bz}{\mathbf{z}}

\newcommand{\bL}{\mathbf{L}}

\newcommand{\bI}{\mathbf{I}}

\newcommand{\bW}{\mathbf{W}}
\newcommand{\bB}{\mathbf{B}}

\newcommand{\bmu}{\boldsymbol{\mu}}

\newcommand{\btheta}{\boldsymbol{\theta}}
\newcommand{\bphi}{\boldsymbol{\phi}}
\newcommand{\bSigma}{\mathbf{\Sigma}}
\newcommand{\bsigma}{\boldsymbol{\sigma}}
\newcommand{\bepsilon}{\boldsymbol{\epsilon}}

\newcommand{\vbar}{\,|\,}

\newcommand{\bzlr}{( \bz )}

\newcommand{\bmuz}{\bmu \bzlr}

\newcommand{\brhox}{\boldsymbol{\rho} ( \bx )}
\newcommand{\bomegax}{\boldsymbol{\omega} ( \bx )}
\newcommand{\bSigmaz}{\bSigma\bzlr}

\newcommand{\bsigmaz}{\bsigma \bzlr}
\newcommand{\bsigmazu}[1]{\bsigma_{#1} \bzlr}
\newcommand{\bmuzu}[1]{\bmu_{#1} \bzlr}

\newcommand{\transpose}{^{\mathtt{T}}}

\newcommand{\NormalDistrib}[1]{\mathcal{N}\big(\, #1 \,\big)}

\newcommand{\bLambda}{\mathbf{\Lambda}}

\newcolumntype{Y}{>{\centering\arraybackslash}X}
\newcolumntype{C}[1]{>{\centering\let\newline\\\arraybackslash\hspace{0pt}}m{#1}}

\usepackage{xspace}
\makeatletter
\DeclareRobustCommand\onedot{\futurelet\@let@token\@onedot}
\def\@onedot{\ifx\@let@token.\else.\null\fi\xspace}
\def\eg{\emph{e.g}\onedot}

\makeatother

\title{Training VAEs Under Structured Residuals} 

\author{Garoe Dorta$^{1,2}$ \enspace  Sara Vicente$^{2}$ \enspace Lourdes Agapito$^{3}$ \enspace Neill D.F. Campbell$^{1}$ \enspace Ivor Simpson$^{2}$\\
$^{1}$University of Bath \qquad $^{2}$Anthropics Technology Ltd. \qquad $^{3}$University College London \\
{\tt \scriptsize $^{1}$\{g.dorta.perez,n.campbell\}@bath.ac.uk $^{2}$\{sara,ivor\}@anthropics.com $^{3}$l.agapito@cs.ucl.ac.uk}}

\begin{document}

\maketitle

\begin{abstract}
Variational auto-encoders (VAEs) are a popular and powerful deep generative model.
Previous works on VAEs have assumed a factorized likelihood model, whereby the output uncertainty of each pixel is assumed to be independent.
This approximation is clearly limited as demonstrated by observing a residual image from a VAE reconstruction, which often possess a high level of structure.
This paper demonstrates a novel scheme to incorporate a structured Gaussian likelihood prediction network within the VAE that allows the residual correlations to be modeled.
Our novel architecture, with minimal increase in complexity, incorporates the covariance matrix prediction within the VAE. We also propose a new mechanism for allowing structured uncertainty on color images.
Furthermore, we provide a scheme for effectively training this model, and include some suggestions for improving performance in terms of efficiency or modeling longer range correlations.
\end{abstract}

\def\plotw{0.13\linewidth}

\def\1img{3}

\begin{figure}[h!]
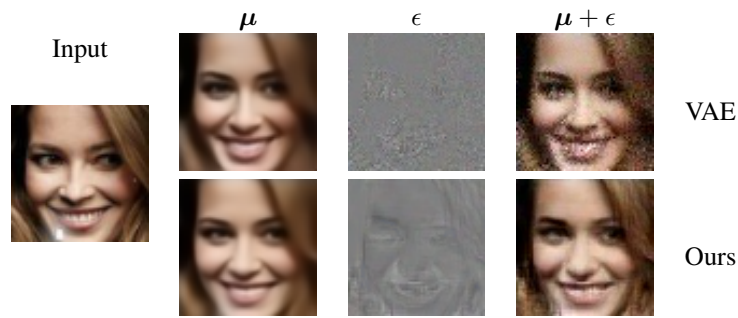

	\centering
	\begin{tabular}{ccccccc}
 	\multirow{1}{*}[-1.2em]{Input} & $\bmu$ & $\epsilon$ & $\bmu + \epsilon$ &  \\
	\multirow{2}{*}[1.7em]{\includegraphics[width=\plotw]{img/celeba/teaser_fig/input/\1img}} &
	\includegraphics[width=\plotw]{img/celeba/teaser_fig/diag/recons/\1img} &
	\includegraphics[width=\plotw]{img/celeba/teaser_fig/diag/residuals/\1img} &
	\includegraphics[width=\plotw]{img/celeba/teaser_fig/diag/recons_with_residual/\1img} & 
	\multirow{1}{*}[1.9em]{VAE} \\
	&
	\includegraphics[width=\plotw]{img/celeba/teaser_fig/covar/recons/\1img} &
	\includegraphics[width=\plotw]{img/celeba/teaser_fig/covar/residual/\1img} &
	\includegraphics[width=\plotw]{img/celeba/teaser_fig/covar/recons_with_residual/\1img} &
	\multirow{1}{*}[1.9em]{Ours}
	\end{tabular}
	\caption{Given an input image, reconstructions from a VAE and our model are shown.
	The VAE models the output distribution as a factorized Gaussian, while our model uses a structured Gaussian likelihood.
	We show the means $\bmu$ and a sample $\bepsilon$ from the corresponding covariances.
	The correlated noise sample of our model better captures the structure present in natural images.}
	\label{fig:teaser}
\end{figure}


\section{Introduction}
\label{sec:introduction}
Generative probabilistic models are a popular tool for estimating data density distributions of images.
Aside from reconstruction and interpolation, deep generative models allow for synthesizing novel examples from the learned distribution.
There are a number of interesting applications for such models; for example, image super-resolution~\cite{Ledig2017SRGAN}, editing images based on attributes~\cite{Yan2016VAEattr2img}, disentangling shading and albedo~\cite{Shu2017NeuralEdit} or noise removal~\cite{Vincent2008DAE}. 

We are interested in generative models with explicit likelihood functions.
These types of models are, in general, sufficiently flexible to learn complex distributions and
are efficient at inference and learning.
They can also be well suited to reconstructing images, which is useful in certain applications.

Variational AutoEncoders (VAEs)~\cite{Kingma2014VAE,Rezende2014VAE_DGLM} are a powerful family of deep generative models that perform variational inference in order to learn high-dimensional distributions on large datasets.
In this model, a mapping is learned from a latent representation to image space using a set of training examples. 
To be able to compute the variational expressions, the associated distributions for both the latent parameters and the residual distribution must have explicit analytic forms.
In general, factorized Gaussian likelihoods are the usual choice for the distributions due to their simplicity.

For image data, factorized likelihoods imply that the residual error at each pixel is distributed independently.
In contrast, correlated likelihood models can account for some spatial structure in the uncertainty distribution.
The factorized assumption is demonstrably false in practical applications, where the residual image often exhibits a clear spatial structure.
The difference between factorized and correlated likelihoods can be highlighted by comparing samples from the output of uncertainty distributions, as shown in Fig.~\ref{fig:teaser}\,.

Recent work~\cite{Dorta2018CVPR} demonstrated that it is possible to predict the structured output uncertainty of a generated image given a learned latent representation.
In essence estimating a covariance matrix from a single sample.
Given the clear limitations of the factorized model, we hypothesize that modeling the structure of residuals should be beneficial to training VAEs.
To investigate this, we propose to extend the VAE by using a structured Gaussian likelihood distribution. 

Providing the VAE with a structured noise model endows it with the capability of modeling complex, high-frequency features, which do not reliably occur in the same location (\eg hair), stochastically rather than deterministically.
This allows the VAE to concentrate on structure that it can model correctly.


To the best of our knowledge, this work shows  the first approach for training VAEs with a correlated Gaussian likelihood.
A naive approach would introduce $(n_p^2 - 1)/2 + n_p$ parameters, where $n_p$ is the number of pixels in the image.
This is infeasible with standard strategies for training deep neural networks.

In this paper we show how to efficiently overcome these limitations.
We have three main contributions.
(1) Providing a novel architecture that combines the VAE and structured covariance prediction models, while limiting the number of additional parameters over the original VAE.
The proposed architecture also allows for structured uncertainty prediction in color images.
(2) An investigation into effective training strategies to avoid poor local minima.
(3) Enhancements to the covariance prediction model of~\cite{Dorta2018CVPR} for improved efficiency (particularly for high dimensional data) and allowing longer range correlations to be modeled.
We show experiments on the CelebA and LSUN Churches~\cite{Yu2015Lsun} datasets and demonstrate that, in terms of the predicted residual distribution, our model offers significant improvements over the factorized Gaussian VAE.  


\section{Related work}
\label{sec:related_work}

This work considers VAE~\cite{Kingma2014VAE}, which employs a maximum likelihood approach, with a variational approximation on the posterior distribution.
VAEs commonly use factorized Gaussian likelihoods for modeling image data.
Previous work can be classified into: methods that improve the flexibility of the posterior distribution, and methods that improve the flexibility of the likelihood distribution.

The majority of recent research into VAEs has been focused on improving the flexibility of the posterior latent distribution.
Such methods include: parametrized transformations of the distribution with tractable Jacobians~\cite{Rezende2015NF,Kingma2016IAF}, 
and employing adversarial training~\cite{Goodfellow2014GAN}, distributions without tractable probability density functions can be adopted~\cite{Makhzani2016AdversarialAE, Mescheder2017AVB}.
Implicit representations via multiple sampling~\cite{Burda2016IWAE} and  Stein particle-based representations~\cite{Pu2017Stein} have also been explored.
This approaches seek to reduce the error induced by the variational approximation on the posterior distribution.
However, they still employ the simple factorized assumption on the residual distribution.
They are orthogonal to our work, and any of them could in principle be used in conjunction with our correlated Gaussian likelihood.

More related to our work are methods that improve the flexibility of the likelihood distribution.
An implicit representation based on Generative Adversarial Networks (GANs)~\cite{Goodfellow2014GAN} has been explored~\cite{Larsen2015VAEGAN}.
However, this method suffers from known issues in GAN training, such as mode collapse and unstable training~\cite{Radford2016DCGAN}.
Much work~\cite{Berthelot2017BEGAN,Arjovsky2017WGAN} has been done to attempt to address those issues with GANs models, yet a satisfactory solution is still to be found.
Autoregressive distributions allow for complex correlations in the data to be modeled.
PixelCNN~\cite{Oord2016PixelRNN} offers a tractable neural approximation to learn these distributions.
Previous work has used PixelCNN distributions for VAEs~\cite{Chen2017LossyVAE,Gulrajani2017PixelVAE,Makhzani2017PixelGAN_AE}, achieving impressive results.
The sequential nature of the distribution, requires a forward pass of the network for each pixel that is generated, which makes it computationally demanding.
In contrast our method is able to model correlations and generate samples in a single forward pass of the network.

Most work on structured uncertainty is only applicable to restricted settings, such as small data scenarios~\cite{nikias1988}, temporally correlated noise models~\cite{WOOLRICH2001} and in Gaussian processes~\cite{rasmussen2006}.
However these techniques in general do not scale well to deep learning settings which typically use large quantities of high-dimensional data.
Methods applicable to deep learning include factorized Gaussian heteroscedastic noise prediction in image segmentation networks~\cite{kendall2017}, and correlated Gaussian noise prediction from a latent image representation~\cite{Dorta2018CVPR}.

\section{Variational Autoencoder}
\label{sec:vae}

We start by reviewing the VAE~\cite{Kingma2014VAE} model.
A VAE consists of (i) a decoder, $p_{\btheta}(\bx \vbar \bz)$, which models the probability distribution of the input data, $\bx$, conditioned on a low-dimensional representation,
$\bz$, in a latent space and (ii) an encoder, $q_{\bphi}(\bz \vbar \bx)$, which models the reverse process.
Neural networks parametrized by $\btheta$ and $\bphi$, are used to model the distributions.

The parameters of the neural networks are estimated such that the marginal likelihood of the input data, $p_{\btheta}(\bx)$, is maximized under a variational Bayesian approximation:
\begin{equation}
\log p_{\btheta}(\bx) = D_{KL}[q_{\bphi}(\bz \vbar \bx)||p_{\btheta}(\bz \vbar \bx)] + L_{VAE},
\label{eq:vae-untractable}
\end{equation}
where the variational lower bound is
\begin{equation}
L_{\text{VAE}} = \mathbb{E}_{\bz \sim q_{\bphi}(\bz \vbar \bx)} \left[ \log ~ p_{\btheta}(\bx \vbar \bz) \right]
 - D_{KL}\left[q_{\bphi}(\bz \vbar \bx)|| p_{\btheta}(\bz)\right].
\label{eq:vae-model}
\end{equation}

In the right-hand side of Eq.~\ref{eq:vae-untractable}, the first term measures the distance between the approximate and the true unknown posterior.
In the right-hand side of the variational lower bound, the first term is the reconstruction error, and the second term is the KL divergence between the encoder distribution and a known prior.
Maximizing the bound will be approximately equivalent to maximizing the marginal likelihood, as long as the approximate posterior distribution is flexible enough.

For continuous data, the approximate posterior and the data likelihood usually take the form of multivariate Gaussian distributions with factorized covariance matrices
\begin{align}
 q_{\bphi}(\bz \vbar \bx) &= \NormalDistrib{\brhox, \bomegax^2 \, \bI}, \\
 p_{\btheta}(\bx \vbar \bz) &= \NormalDistrib{\bmuz, \bsigmaz^2 \, \bI},
 \label{eq:generative_model}
\end{align}
where $\bx$ is the image as a column vector, the means $\bmuz, \brhox$ and variances $\bsigmaz^2, \bomegax^2$ are (non-linear) functions of the inputs or the latent variables.

%
%
\section{Methodology}
\label{sec:methodology}
We propose to extend the VAE to use a correlated Gaussian likelihood
\begin{align}
 p_{\btheta}(\bx \,|\, \bz) &= \NormalDistrib{\bmuz, \bSigmaz},
 \label{eq:generative_model_cov}
\end{align}
where $\bSigmaz$ is a dense covariance matrix.
The covariance matrix captures the correlations between pixels, thus allowing the model to predict correlated uncertainty over its outputs.

The learning task for the model is ill-posed, as a full covariance matrix $\bSigmaz$ must be predicted from each input using its encoded latent representation.
Moreover, the initial estimates are far away from any reasonable solution, as the weighs of the networks are initialized randomly.

The first issue can be partly overcome by leveraging previous work on structured uncertainty prediction for deep neural networks~\cite{Dorta2018CVPR}.
In that work, the authors tackled the problem by restricting the uncertainty prediction network to only be able to model covariance matrices that have sparse inverses.
Formally, these restricted covariance matrices are defined as 
\begin{equation}
\bSigmaz^{-1} = \bLambda(\bz)= \bL(\bz) \bL(\bz)\transpose,
\end{equation}
where the $\bLambda(\bz)$ is a sparse precision matrix, and $\bL(\bz) \bL(\bz)\transpose$ is its Cholesky decomposition.

The advantages of modeling the covariance in this way, is that the uncertainty prediction network is only required to estimate the non-zero values in $\bL(\bz)$, and it is trivial to evaluate all the terms of a Gaussian likelihood from $\bL(\bz)$.
Moreover, despite $\bLambda(\bz)$ being sparse, $\bSigmaz$ remains a dense matrix, which allows modeling long range correlations in the residuals.
The sparsity pattern proposed by the authors is such that, for a predefined patch size $n_{\text{f}}$, pixels that are inside the  $n_{\text{f}}$-neighborhood in image space have non-zero entries in $\bL(\bz)$.
For an input image with $n_{\text{p}}$ pixels there are $n_{\text{p}} \times (n_{\text{f}}^2 - 1)/2 + 1$ non-zero entries in $\bL(\bz)$, as $\bL(\bz)$ is a lower triangular matrix.
The number of parameters of the covariance prediction network is proportional to the square of the neighborhood size $n_{\text{f}}$. 


The $\bL(\bz)$ network as described in~\cite{Dorta2018CVPR}, can only be used to estimate sparse precision matrices for small gray-scale images.
In the following sections we will show how to apply the method within the context of VAEs, and how to handle color images as well as larger resolution inputs.

\subsection{Color images}
We present a structure uncertainty approach that can model color images with a minimal increment in the number of parameters over modeling gray-scale images.
To achieve this, we first observe that in a luminance color space, such as YCbCr, the high-frequency details of the image are mostly encoded in the luminance channel.
This fact has been used by image compression algorithms like JPEG, where the color channels Cb and Cr are quantized with minimal loss of quality in the resulting images.
It is known that VAEs struggle to model high-frequency details, leading to highly structured residuals for the luminance channel, in contrast the Cb and Cr, which are smooth by nature, lead to mostly uncorrelated residuals, as shown in Fig.~\ref{fig:ycbcr_residuals}.

\def\plotwYCbCResidual{0.11\linewidth}
\begin{figure}[t!]
	\centering
	\setlength{\tabcolsep}{1pt} 
	\begin{tabularx}{\linewidth}{ccccXcccc}
	& {\small Input } & {\small Reconstruction} & {\small Residual} & & & {\small Input } & {\small Reconstruction} & {\small Residual}\\
	Y \quad & \includegraphics[align=c,width=\plotwYCbCResidual]{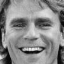} &
	\includegraphics[align=c,width=\plotwYCbCResidual]{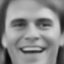} &
	\includegraphics[align=c,width=\plotwYCbCResidual]{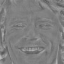} &
	& R \quad & \includegraphics[align=c,width=\plotwYCbCResidual]{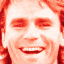} &
	\includegraphics[align=c,width=\plotwYCbCResidual]{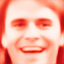} &
	\includegraphics[align=c,width=\plotwYCbCResidual]{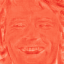}  \\[23pt]
	Cb \quad & \includegraphics[align=c,width=\plotwYCbCResidual]{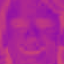} &
	\includegraphics[align=c,width=\plotwYCbCResidual]{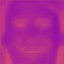} &
	\includegraphics[align=c,width=\plotwYCbCResidual]{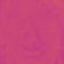} &
	& G \quad& \includegraphics[align=c,width=\plotwYCbCResidual]{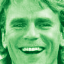} &
	\includegraphics[align=c,width=\plotwYCbCResidual]{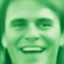} &
	\includegraphics[align=c,width=\plotwYCbCResidual]{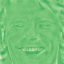} \\[23pt]
	Cr \quad &	\includegraphics[align=c,width=\plotwYCbCResidual]{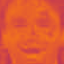} &
	\includegraphics[align=c,width=\plotwYCbCResidual]{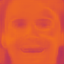} &
	\includegraphics[align=c,width=\plotwYCbCResidual]{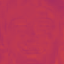} &
	& B \quad &	\includegraphics[align=c,width=\plotwYCbCResidual]{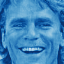} &
	\includegraphics[align=c,width=\plotwYCbCResidual]{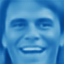} &
	\includegraphics[align=c,width=\plotwYCbCResidual]{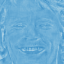}
	\end{tabularx}
	\caption{Input, reconstructions and residuals in YCbCr and RGB color spaces for a VAE with diagonal covariance trained with RGB images.
	In the YCbCr space the residuals of the Y channel are highly structured, while the ones for the color channels are not.
	In RGB space all the channels contain highly structured residuals and the information is highly correlated between the channels.}
	\label{fig:ycbcr_residuals}
\end{figure}

Therefore, the luminance channel is modeled using a correlated Gaussian distribution, while the remaining channels will use factorized Gaussians
\begin{align}
 \bx &= [\bx_Y, \bx_{Cb}, \bx_{Cr}], \\
 p_{\btheta}(\bx \,|\, \bz) &=  p_{\btheta}(\bx_{Y} \,|\, \bz)  p_{\btheta}(\bx_{Cb} \,|\, \bz)  p_{\btheta}(\bx_{Cr} \,|\, \bz), \\
 p_{\btheta}(\bx_Y \,|\, \bz) &= \NormalDistrib{\bmuzu{Y}, \bSigmaz}, \\
 p_{\btheta}(\bx_{Cb} \,|\, \bz) &= \NormalDistrib{\bmuzu{Cb}, \bsigmazu{Cb}^2 \bI}, \\
 p_{\btheta}(\bx_{Cr} \,|\, \bz) &= \NormalDistrib{\bmuzu{Cr}, \bsigmazu{Cr}^2 \bI}.
 \label{eq:generative_model_ycbcr}
\end{align}


Many datasets contain images compressed in JPEG, and as aforementioned this format quantizes the Cb Cr channels.
The loss of information due to quantization can be problematic, as with different amount of information the color channels should be treated differently.
In order to equalize the amount of information per pixel across channels, the Cb Cr channels are downsampled with a factor proportional to the quantization factor. 

\subsection{Priors}
As our generative model estimates both a mean and a covariance matrix per input image, we must employ some regularization to avoid poor solutions.
Intuitively, little variance is desired on the predicted covariance matrix, i.e. the model should be certain about its predictions.
Another issue is the prediction of spurious correlations, which are not well supported by the data.
Experimentally, we find that without any regularization a randomly initialized network has a tendency to model most of the information in the covariance matrix $\bSigmaz$, which is undesirable.

The Bayesian approach for regularization is to add a prior distribution over the predicted parameters.
The standard prior used for covariance matrices is an inverse Wishart distribution, as this is a conjugate prior for the normal distribution.
Alternatively, we could consider a Gamma distribution for the diagonal values in $\bL(\bz)$ and a Gaussian for the off-diagonal ones.

In practice, finding a good set of parameters for the priors that encouraged low variance and sparse correlation structure was difficult to find.
The total likelihood of the model was in dominated by the cost of the priors. 
Both the Wishart and the Gamma-Gaussian priors made the posterior too strongly diagonal.
Empirically, we obtained good results with a variance minimizer and an $L_1$ regularizer for the off-diagonal elements in $\bL(\bz)$, where the loss is defined as
\begin{equation}
L = L_{\text{VAE}} + \alpha ||\bsigmaz^2||_1 + \gamma \sum_{i,j} | \bL(\bz)_{i,j} |,
\end{equation}
where $\alpha$ and $\gamma$ are scalar hyper-parameters and $i \neq j$.
Evaluating the estimated variance $\bsigmaz^2$ given $\bL(\bz)$ implies performing a costly matrix inverse.
Instead we approximate the term using the empirical variance $||\bsigmaz^2||_1 \approx || \bx - \bmuz ||_2^2$.

\subsection{Scaling to larger images}
To model larger images, the size of the neighborhood should be increased accordingly, so that relevant correlations are still modeled.
However, the dimensionality of the covariance matrix increases quadratically with the size of the neighborhood.
Our proposed solution is to reduce the dimensionality of $\bL(\bz)$ by approximating it with a learnable basis
\begin{equation}
\bL(\bz) = s(\bB \bW(\bz)),
\label{eq:L_with_basis}
\end{equation}
where $\bB$ is a $(n_{\text{f}}^2 - 1)/2 + 1 \times n_{\text{b}}$ matrix containing the basis, $\bW(\bz)$ is a $n_{\text{b}} \times n_{\text{p}}$ matrix of weights, $n_{\text{p}}$ is the number of pixels in the input, $n_{\text{b}}$ is the number of basis vectors and $n_{\text{f}}$ is the neighborhood size.
Each column in the matrix $\bB \bW(\bz)$ contains a dense representation of the corresponding column in $\bL$, and the operator $s(\cdot)$ pads with zeroes its input, converting from the dense representation of $\bB \bW(\bz)$ to the sparse one of $\bL(\bz)$.

The covariance network output becomes $\bW(\bz)$, while the basis $\bB$ is learned at train time and it is shared for all the images.
To further boost the reduction in dimensionality in $\bL(\bz)$, the bases $\bB$ can be constructed such that the neighboring structure is similar to dilated convolutions. 

Experimentally, we saw only marginal gains when using the basis, the dilated sparsity pattern and both together.
However, dilated convolutions have been shown to be a good approximation to large dense filters.
Therefore, we believe such dilated-like sparsity patterns might be useful for larger resolution images.


\subsection{Efficiency}

Our approach can be implemented efficiently in modern GPU architectures.
During learning the conditional log likelihood of the data is evaluated as
\begin{equation}
\log p_{\theta}(\bx \vert \bz) = - \frac{1}{2} \left( \log | \bLambda(\bz) | + (\bx - \bmuz)^{\transpose} \bLambda(\bz) (\bx - \bmuz) + n_p \log(2 \pi) \right).
\end{equation}

The basis derivation in the previous section is used, where $\bL(\bz) = s(\bB \bW(\bz))$.
The square error term can be evaluated as $\by \by^{\transpose}$, where $\by = (\bx - \bmu)\transpose \bL(\bz)$,  which avoids explicitly constructing $\bLambda$.
Each column in the matrix $\bB$ is reshaped and zero padded to an $n_\text{f} \times n_\text{f}$ kernel.
The squared error is computed by convolving the residual with each kernel, and performing a linear combination with the weights $\bW(\bz)$.
The log determinant term can be evaluated as $2 \sum_i\log \left( \bB_{0,i} \bW(\bz) \right)$, where $\bB_{0,i}$ is a vector.
If no basis matrix is used, the same convolutional approach can be applied by setting $\bB = \bI$.
During inference, the sampling approach described in~\cite{Dorta2018CVPR} is used.



\section{Results}
\label{sec:results}

We evaluate our model on the CelebA~\cite{Liu2015CelebADataset} and LSUN Outdoor Churches~\cite{Yu2015Lsun} datasets.
The networks are implemented in Tensorflow~\cite{Tensorflow} and they are trained on a single Titan X GPU using the Adam~\cite{Kingma2015Adam} optimizer.
All experiments use images of $64 \times 64$ pixels, where the Cb and Cr channels are blurred and downsampled to $16 \times 16$ pixels.
The patch size $n_f$ for our covariance prediction is set to 3.
Additional details on the model architecture can be found in the supplemental material.
For data augmentation we employ simple left-right flips of the images.

All models are trained with images on the YCbCr space, which allows a direct comparison.
Two factorizations are tested for VAEs, a spherical covariance for both the Y and Cb and Cr channels, denoted as VAE (Sph), and a diagonal covariance for the Y channel and spherical for Cb and Cr, denoted VAE (Diag).
Results for VAEs trained with RGB data are shown in the supplement.

A batch size of $64$ and a learning rate of $0.0005$ is used.
VAE (Sph) is used as pretraining for our model, additionally for the first $5$ epochs only the structured uncertainty branch of the network is trained.
The hyper-parameters values for the regularizers are $\alpha = 10$ and $\beta = 0.001$.

\subsection{CelebA}

We use the aligned and cropped version of the dataset, and we further crop the image in a square centered at the face, following the same procedure as~\cite{Larsen2015VAEGAN}.
All the models are trained for 110 epochs.

VAEs with factorized Gaussian likelihoods can overfit to the reconstruction error, thus producing low quality samples.
$\beta$-VAE~\cite{Higgins2017BVAE} address this issue by increasing the weight of the KL term in the log likelihood.
We show results for both a VAE and $\beta$-VAE trained with a diagonal Gaussian likelihood.
For $\beta$-VAE the authors recommend a value of $\beta$ of $62.5$ for this dataset, however we found experimentally that a value of $5$ performed better.
For a fair comparison with SUPN~\cite{Dorta2018CVPR}, the pretrained VAE that is needed for the method is trained using our approach.



Quantitative results are shown in Table~\ref{tb:celeba_error_table}~.
The lower bound of the negative log likelihood on the test set, evaluated with 500 $\bz$ samples per image as described in~\cite{Burda2016IWAE}.
For $\beta$-VAE we report the likelihood after setting $\beta=1$.
Values are reported for both grayscale and YCbCr images.
Our method achieves significantly lower likelihood than competing methods, with the exception of~\cite{Dorta2018CVPR}.
However, our model is less complex as we do not require a separate decoder network for the structured covariance prediction.



The KL divergence between the approximate posterior and the prior show that our correlated residual distribution prediction is also beneficial in order to achieve a latent distribution that better follows the prior.

\begin{table}[t!]
\centering
\setlength{\tabcolsep}{2pt}
\tiny
\begin{tabularx}{\linewidth}{YYYYYY} 
	\small \textbf{Model} & \small \textbf{NLL} & \small \textbf{KL} & \small \textbf{FLOPs} \\ \midrule 
	\small VAE (Sph)~\cite{Kingma2014VAE}  & $-3647 \pm 1033 ~/~ -4517 \pm 1308$ & $326.31 ~/~ 322.94$ & $6.67e6 ~/~ 6.68e6$  \\ \midrule
	\small VAE (Diag)~\cite{Kingma2014VAE}  & $-4016 \pm \phantom{0}813 ~/~ -4598 \pm 2747$ & $339.69 ~/~ 341.92$ & $6.66e6 ~/~ 6.67e6$  \\ \midrule
	\small $\beta$-VAE (Diag)~\cite{Higgins2017BVAE}  & $-3123 \pm \phantom{0}998 ~/~ -5574 \pm 1083$ & $\phantom{0}42.48 ~/~ 199.16$ & $6.66e6 ~/~ 6.67e6$ \\ \midrule
	\small  SUPN~\cite{Dorta2018CVPR} & $-8308 \pm 1455 ~/~ \phantom{000000}-\phantom{0000}$ & $ 269.76 ~/~ \phantom{0}-\phantom{00}$ & $9.78e6 ~/~ \phantom{0}-\phantom{00}$  \\ \midrule
	\small Ours & $-8297 \pm 1455 ~/~ -8669 \pm 1424$ & $269.76 ~/~ 281.91$ & $6.72e6 ~/~ 6.70e6$ \\ \bottomrule
\end{tabularx} \vspace{1pt}
\caption{Quantitative comparison of density estimation error measured as the negative log likelihood (NLL) for the (grayscale / YCbCr) CelebA dataset, lower is better.
FLOPs measures models complexity, and KL denotes the KL divergence of the approximate posterior to the prior.
Our model is able to achieve a likelihood similar to previous work~\cite{Dorta2018CVPR}, with a significant reduction in model complexity.
}
\label{tb:celeba_error_table}
\end{table}

Reconstructions are shown in Fig.~\ref{fig:celeba_reconstructions}.
The means, $\bmu$, obtained with our model are comparable to competing methods and the samples, $\bepsilon$, from the noise model add plausible details, like hair.

\def\plotw{0.1\linewidth}

\def\1img{9}
\def\2img{11}
\def\3img{3}
\def\4img{13}

\begin{figure}[t!]
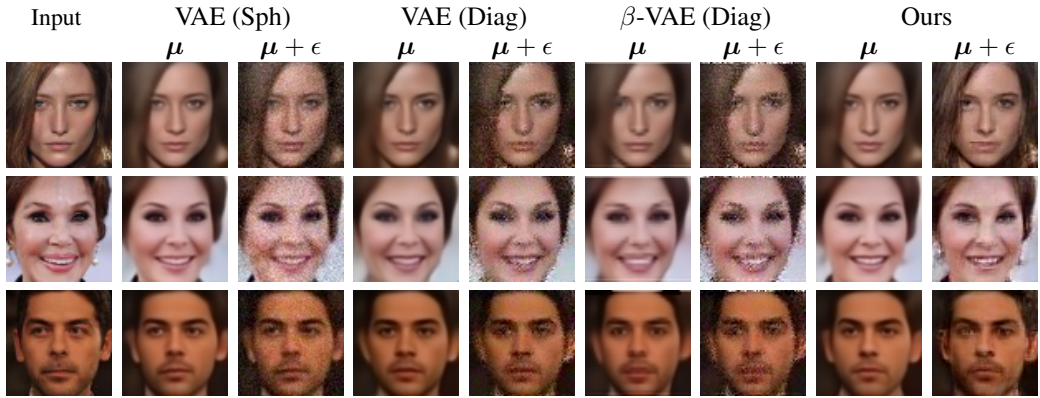

	\centering
	\setlength{\tabcolsep}{2pt} 
	\begin{tabular}{ccccccccc}
	{\small Input } & \multicolumn{2}{c}{VAE (Sph)} & \multicolumn{2}{c}{VAE (Diag)} & \multicolumn{2}{c}{$\beta$-VAE (Diag)} & \multicolumn{2}{c}{Ours} \\
	 & $\bmu$ & $\bmu + \epsilon$ & $\bmu$ & $\bmu + \epsilon$ & $\bmu$ & $\bmu + \epsilon$ & $\bmu$ & $\bmu + \epsilon$ \\
	\includegraphics[width=\plotw]{img/celeba/recons_fig_rgb/input/\1img} &
	\includegraphics[width=\plotw]{img/celeba/recons_fig_ycbcr/sph/recons/\1img} &
	\includegraphics[width=\plotw]{img/celeba/recons_fig_ycbcr/sph/recons_with_residual/\1img} &
	\includegraphics[width=\plotw]{img/celeba/recons_fig_ycbcr/diag_sph/recons/\1img} &
	\includegraphics[width=\plotw]{img/celeba/recons_fig_ycbcr/diag_sph/recons_with_residual/\1img} &
	\includegraphics[width=\plotw]{img/celeba/recons_fig_ycbcr/diag_sph_beta_1_5/recons/\1img} &
	\includegraphics[width=\plotw]{img/celeba/recons_fig_ycbcr/diag_sph_beta_1_5/recons_with_residual/\1img} &
	\includegraphics[width=\plotw]{img/celeba/recons_fig_ycbcr/covar/recons/\1img} &
	\includegraphics[width=\plotw]{img/celeba/recons_fig_ycbcr/covar/recons_with_residual/\1img} \\
	\includegraphics[width=\plotw]{img/celeba/recons_fig_rgb/input/\2img} &
	\includegraphics[width=\plotw]{img/celeba/recons_fig_ycbcr/sph/recons/\2img} &
	\includegraphics[width=\plotw]{img/celeba/recons_fig_ycbcr/sph/recons_with_residual/\2img} &
	\includegraphics[width=\plotw]{img/celeba/recons_fig_ycbcr/diag_sph/recons/\2img} &
	\includegraphics[width=\plotw]{img/celeba/recons_fig_ycbcr/diag_sph/recons_with_residual/\2img} &
	\includegraphics[width=\plotw]{img/celeba/recons_fig_ycbcr/diag_sph_beta_1_5/recons/\2img} &
	\includegraphics[width=\plotw]{img/celeba/recons_fig_ycbcr/diag_sph_beta_1_5/recons_with_residual/\2img} &
	\includegraphics[width=\plotw]{img/celeba/recons_fig_ycbcr/covar/recons/\2img} &
	\includegraphics[width=\plotw]{img/celeba/recons_fig_ycbcr/covar/recons_with_residual/\2img} \\
	\includegraphics[width=\plotw]{img/celeba/recons_fig_rgb/input/\3img} &
	\includegraphics[width=\plotw]{img/celeba/recons_fig_ycbcr/sph/recons/\3img} &
	\includegraphics[width=\plotw]{img/celeba/recons_fig_ycbcr/sph/recons_with_residual/\3img} &
	\includegraphics[width=\plotw]{img/celeba/recons_fig_ycbcr/diag_sph/recons/\3img} &
	\includegraphics[width=\plotw]{img/celeba/recons_fig_ycbcr/diag_sph/recons_with_residual/\3img} &
	\includegraphics[width=\plotw]{img/celeba/recons_fig_ycbcr/diag_sph_beta_1_5/recons/\3img} &
	\includegraphics[width=\plotw]{img/celeba/recons_fig_ycbcr/diag_sph_beta_1_5/recons_with_residual/\3img} &
	\includegraphics[width=\plotw]{img/celeba/recons_fig_ycbcr/covar/recons/\3img} &
	\includegraphics[width=\plotw]{img/celeba/recons_fig_ycbcr/covar/recons_with_residual/\3img}
	\end{tabular}
	\caption{Comparison of image reconstructions for the different models.
	In contrast to previous work, our model is able to learn structured residuals.}
	\label{fig:celeba_reconstructions}
\end{figure}

Samples of all the models are shown in Fig.~\ref{fig:celeba_samples}.
The VAE is over confident in its predictions, as denoted by the high value in the KL divergence shown in Table.~\ref{tb:celeba_error_table}.
Consequently, this leads to a latent space that does not follow the prior distribution and thus to the poor samples observed.
$\beta$-VAE is able to produce samples that are of similar quality to its reconstructions.
Our method is able to produce good quality samples, where the structured uncertainty prediction branch is again able to model high frequency details.

\def\plotw{0.1\linewidth}

\def\1img{1}
\def\2img{9}
\def\3img{14}

\begin{figure}[t!]
	\centering
	\setlength{\tabcolsep}{2pt} 
	\begin{tabular}{cccccccc}
	\multicolumn{2}{c}{VAE (Sph)} & \multicolumn{2}{c}{VAE (Diag)} & \multicolumn{2}{c}{$\beta$-VAE (Diag)} & \multicolumn{2}{c}{Ours} \\
	$\bmu$ & $\bmu + \epsilon$ &  $\bmu$ & $\bmu + \epsilon$ & $\bmu$ & $\bmu + \epsilon$ & $\bmu$ & $\bmu + \epsilon$ \\
	\includegraphics[width=\plotw]{img/celeba/samples_fig_ycbcr/sph/sample/\1img} &
	\includegraphics[width=\plotw]{img/celeba/samples_fig_ycbcr/sph/sample_with_residual/\1img} &
	\includegraphics[width=\plotw]{img/celeba/samples_fig_ycbcr/diag_sph/sample/\1img} &
	\includegraphics[width=\plotw]{img/celeba/samples_fig_ycbcr/diag_sph/sample_with_residual/\1img} &
	\includegraphics[width=\plotw]{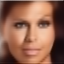} &
	\includegraphics[width=\plotw]{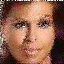} &
	\includegraphics[width=\plotw]{img/celeba/samples_fig_ycbcr/covar/sample/\1img} &
	\includegraphics[width=\plotw]{img/celeba/samples_fig_ycbcr/covar/sample_with_residual/\1img} \\
	\includegraphics[width=\plotw]{img/celeba/samples_fig_ycbcr/sph/sample/\2img} &
	\includegraphics[width=\plotw]{img/celeba/samples_fig_ycbcr/sph/sample_with_residual/\2img} &
	\includegraphics[width=\plotw]{img/celeba/samples_fig_ycbcr/diag_sph/sample/\2img} &
	\includegraphics[width=\plotw]{img/celeba/samples_fig_ycbcr/diag_sph/sample_with_residual/\2img} &
	\includegraphics[width=\plotw]{img/celeba/samples_fig_ycbcr/diag_sph_beta_1_5/sample/\2img} &
	\includegraphics[width=\plotw]{img/celeba/samples_fig_ycbcr/diag_sph_beta_1_5/sample_with_residual/\2img} &
	\includegraphics[width=\plotw]{img/celeba/samples_fig_ycbcr/covar/sample/\2img} &
	\includegraphics[width=\plotw]{img/celeba/samples_fig_ycbcr/covar/sample_with_residual/\2img} \\
	\includegraphics[width=\plotw]{img/celeba/samples_fig_ycbcr/sph/sample/\3img} &
	\includegraphics[width=\plotw]{img/celeba/samples_fig_ycbcr/sph/sample_with_residual/\3img} &
	\includegraphics[width=\plotw]{img/celeba/samples_fig_ycbcr/diag_sph/sample/\3img} &
	\includegraphics[width=\plotw]{img/celeba/samples_fig_ycbcr/diag_sph/sample_with_residual/\3img} &
	\includegraphics[width=\plotw]{img/celeba/samples_fig_ycbcr/diag_sph_beta_1_5/sample/\3img} &
	\includegraphics[width=\plotw]{img/celeba/samples_fig_ycbcr/diag_sph_beta_1_5/sample_with_residual/\3img} &
	\includegraphics[width=\plotw]{img/celeba/samples_fig_ycbcr/covar/sample/\3img} &
	\includegraphics[width=\plotw]{img/celeba/samples_fig_ycbcr/covar/sample_with_residual/\3img}
	\end{tabular}
	\caption{ Samples for all models, where our method is the only able to include high-frequency details such as wrinkles.}
	\label{fig:celeba_samples}
\end{figure}


\subsection{LSUN}

We use the \textit{church outdoors} category of this dataset, as the test data is not available we use the validation set instead.
All the models are trained for 150 epochs.

Quantitative results for reconstructions are presented in Table~\ref{tb:lsun_error_table}, where we measure the mean squared error (MSE) with respect to the input, as well as the negative log likelihood.
The MSE is measured using the RGB images before the conversion to YCbCr color space, and it is computed using only the predicted means, i.e. $\bepsilon$ is set to zero.
The values correspond to images in the $[0, 255]$ range, and we show mean and standard deviations across the dataset.
Our model is able to offer an improvements over a VAE with diagonal Gaussian, while the $\beta$-VAE shows a significant drop in the reconstruction quality.
A VAE with spherical noise is able to achieve the lowest reconstruction cost.
Our model is able to outperform competing methods in terms of marginal log likelihood.

\begin{table}[t!]
\centering
\setlength{\tabcolsep}{2pt}
\small
\begin{tabularx}{\linewidth}{YYYYY} 
	\textbf{Model} & \textbf{NLL} & \textbf{KL} & \textbf{MSE} \\ \midrule 
	\small VAE (Sph)~\cite{Kingma2014VAE}  & $-2440 \pm 1176$ & $303.83$ & $566 \pm 285$  \\ \midrule
	\small VAE (Diag)~\cite{Kingma2014VAE}  & $-4464 \pm 1924$ & $331.06$ & $728 \pm 327$  \\ \midrule
	\small $\beta$-VAE (Diag)~\cite{Higgins2017BVAE}  & $-4213 \pm 2055$ & $205.88$ & $800 \pm 354$ \\ \midrule
	\small Ours & $-6918 \pm 2423$ & $313.92$ & $614 \pm 286$ \\ \bottomrule
\end{tabularx} \vspace{1pt}
\caption{Quantitative comparison of density estimation error measured as the negative log likelihood (NLL) for the LSUN dataset, lower is better.
Our model is able to achieve a significant improvement in likelihood in comparison with VAEs with factorized noise models.
}
\label{tb:lsun_error_table}
\end{table}


%
%





Reconstructions are shown in Fig.~\ref{fig:lsun_reconstructions}, where we find again that our model is able to produce better reconstructions than competing methods, with the exception of a VAE with spherical noise.
However, the residuals modeled by a spherical VAE are quite limited, as they are forced to have high levels of noise of noise throughout the image, including areas which are trivial to model like the sky.
Our structured residuals add fine detail, however as this dataset is more complex than faces all the models struggle to reconstruct the input.

\def\plotw{0.1\linewidth}

\def\1img{15}
\def\2img{14}
\def\3img{3}
\def\4img{13}

\begin{figure}[t!]
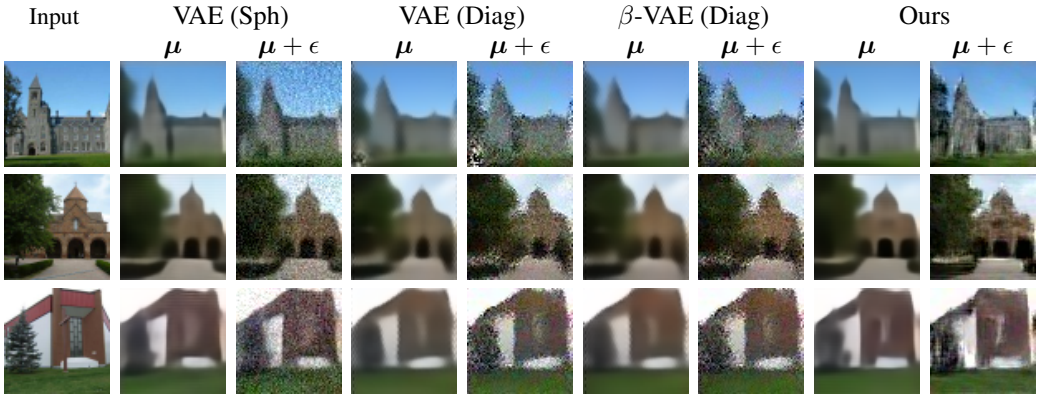

	\centering
	\setlength{\tabcolsep}{2pt} 
	\begin{tabular}{ccccccccc}
	{\small Input } & \multicolumn{2}{c}{VAE (Sph)} & \multicolumn{2}{c}{VAE (Diag)} & \multicolumn{2}{c}{$\beta$-VAE (Diag)} & \multicolumn{2}{c}{Ours} \\
	 & $\bmu$ & $\bmu + \epsilon$ & $\bmu$ & $\bmu + \epsilon$ & $\bmu$ & $\bmu + \epsilon$ & $\bmu$ & $\bmu + \epsilon$  \\
	\includegraphics[width=\plotw]{img/lsun/recons_fig_rgb/input/\1img} &
	\includegraphics[width=\plotw]{img/lsun/recons_fig_ycbcr/sph/recons/\1img} &
	\includegraphics[width=\plotw]{img/lsun/recons_fig_ycbcr/sph/recons_with_residual/\1img} &
	\includegraphics[width=\plotw]{img/lsun/recons_fig_ycbcr/diag_sph/recons/\1img} &
	\includegraphics[width=\plotw]{img/lsun/recons_fig_ycbcr/diag_sph/recons_with_residual/\1img} &
	\includegraphics[width=\plotw]{img/lsun/recons_fig_ycbcr/diag_sph_beta_5/recons/\1img} &
	\includegraphics[width=\plotw]{img/lsun/recons_fig_ycbcr/diag_sph_beta_5/recons_with_residual/\1img} &
	\includegraphics[width=\plotw]{img/lsun/recons_fig_ycbcr/covar/recons/\1img} &
	\includegraphics[width=\plotw]{img/lsun/recons_fig_ycbcr/covar/recons_with_residual/\1img} \\
	\includegraphics[width=\plotw]{img/lsun/recons_fig_rgb/input/\2img} &
	\includegraphics[width=\plotw]{img/lsun/recons_fig_ycbcr/sph/recons/\2img} &
	\includegraphics[width=\plotw]{img/lsun/recons_fig_ycbcr/sph/recons_with_residual/\2img} &
	\includegraphics[width=\plotw]{img/lsun/recons_fig_ycbcr/diag_sph/recons/\2img} &
	\includegraphics[width=\plotw]{img/lsun/recons_fig_ycbcr/diag_sph/recons_with_residual/\2img} &
	\includegraphics[width=\plotw]{img/lsun/recons_fig_ycbcr/diag_sph_beta_5/recons/\2img} &
	\includegraphics[width=\plotw]{img/lsun/recons_fig_ycbcr/diag_sph_beta_5/recons_with_residual/\2img} &
	\includegraphics[width=\plotw]{img/lsun/recons_fig_ycbcr/covar/recons/\2img} &
	\includegraphics[width=\plotw]{img/lsun/recons_fig_ycbcr/covar/recons_with_residual/\2img} \\
	\includegraphics[width=\plotw]{img/lsun/recons_fig_rgb/input/\3img} &
	\includegraphics[width=\plotw]{img/lsun/recons_fig_ycbcr/sph/recons/\3img} &
	\includegraphics[width=\plotw]{img/lsun/recons_fig_ycbcr/sph/recons_with_residual/\3img} &
	\includegraphics[width=\plotw]{img/lsun/recons_fig_ycbcr/diag_sph/recons/\3img} &
	\includegraphics[width=\plotw]{img/lsun/recons_fig_ycbcr/diag_sph/recons_with_residual/\3img} &
	\includegraphics[width=\plotw]{img/lsun/recons_fig_ycbcr/diag_sph_beta_5/recons/\3img} &
	\includegraphics[width=\plotw]{img/lsun/recons_fig_ycbcr/diag_sph_beta_5/recons_with_residual/\3img} &
	\includegraphics[width=\plotw]{img/lsun/recons_fig_ycbcr/covar/recons/\3img} &
	\includegraphics[width=\plotw]{img/lsun/recons_fig_ycbcr/covar/recons_with_residual/\3img}
	\end{tabular}
	\caption{Comparison of image reconstructions for the different models.
	Our model is able to improve over the oversmoothed images generated by previous work.}
	\label{fig:lsun_reconstructions}
\end{figure}

Samples from the models are shown in Fig.~\ref{fig:lsun_samples}, where we find that the VAE with diagonal covariance struggles to generate anything meaningful.
$\beta$-VAE is able to generate recognizable shapes, which are significantly blurry, while our model is able to produce means similar to VAE (Sph) with better noise samples.

\def\plotw{0.1\linewidth}

\def\1img{3}
\def\2img{7}
\def\3img{14}

\begin{figure}[t!]
	\centering
	\setlength{\tabcolsep}{2pt} 
	\begin{tabular}{cccccccc}
	\multicolumn{2}{c}{VAE (Sph)} & \multicolumn{2}{c}{VAE (Diag)} & \multicolumn{2}{c}{$\beta$-VAE (Diag)} & \multicolumn{2}{c}{Ours} \\
	$\bmu$ & $\bmu + \epsilon$ & $\bmu$ & $\bmu + \epsilon$ & $\bmu$ & $\bmu + \epsilon$ & $\bmu$ & $\bmu + \epsilon$  \\
	\includegraphics[width=\plotw]{img/lsun/samples_fig_ycbcr/sph/sample/\1img} &
	\includegraphics[width=\plotw]{img/lsun/samples_fig_ycbcr/sph/sample_with_residual/\1img} &
	\includegraphics[width=\plotw]{img/lsun/samples_fig_ycbcr/diag_sph/sample/\1img} &
	\includegraphics[width=\plotw]{img/lsun/samples_fig_ycbcr/diag_sph/sample_with_residual/\1img} &
	\includegraphics[width=\plotw]{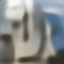} &
	\includegraphics[width=\plotw]{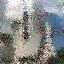} &
	\includegraphics[width=\plotw]{img/lsun/samples_fig_ycbcr/covar/sample/\1img} &
	\includegraphics[width=\plotw]{img/lsun/samples_fig_ycbcr/covar/sample_with_residual/\1img} \\
	\includegraphics[width=\plotw]{img/lsun/samples_fig_ycbcr/sph/sample/\2img} &
	\includegraphics[width=\plotw]{img/lsun/samples_fig_ycbcr/sph/sample_with_residual/\2img} &
	\includegraphics[width=\plotw]{img/lsun/samples_fig_ycbcr/diag_sph/sample/\2img} &
	\includegraphics[width=\plotw]{img/lsun/samples_fig_ycbcr/diag_sph/sample_with_residual/\2img} &
	\includegraphics[width=\plotw]{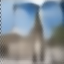} &
	\includegraphics[width=\plotw]{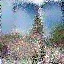} &
	\includegraphics[width=\plotw]{img/lsun/samples_fig_ycbcr/covar/sample/\2img} &
	\includegraphics[width=\plotw]{img/lsun/samples_fig_ycbcr/covar/sample_with_residual/\2img} \\
	\includegraphics[width=\plotw]{img/lsun/samples_fig_ycbcr/sph/sample/\3img} &
	\includegraphics[width=\plotw]{img/lsun/samples_fig_ycbcr/sph/sample_with_residual/\3img} &
	\includegraphics[width=\plotw]{img/lsun/samples_fig_ycbcr/diag_sph/sample/\3img} &
	\includegraphics[width=\plotw]{img/lsun/samples_fig_ycbcr/diag_sph/sample_with_residual/\3img} &
	\includegraphics[width=\plotw]{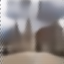} &
	\includegraphics[width=\plotw]{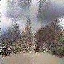} &
	\includegraphics[width=\plotw]{img/lsun/samples_fig_ycbcr/covar/sample/\3img} &
	\includegraphics[width=\plotw]{img/lsun/samples_fig_ycbcr/covar/sample_with_residual/\3img}
	\end{tabular}
	\caption{ Samples drawn from the models.
	VAE fails to learn an useful latent space and $\beta$-VAE produces blurrier images than our model.}
	\label{fig:lsun_samples}
\end{figure}

\section{Conclusions}
This paper proposes the first approach for endowing Variational AutoEncoders with a structured likelihood model.
Our results have demonstrated that VAEs can be successfully trained to predict structured output uncertainty, and that such models have similar reconstructions than those obtained with a factorized likelihood model.
 
In this paper, we have proposed a simple scheme for ensuring the VAE reduces the variance of the model, rather than attempting to describe the residual error through the covariance. This raises interesting avenues for future work, particularly in terms of adding hierarchical priors to prevent the structured residual model adding spurious correlations. Further work will also include investigations on higher resolution images and the efficacy of the basis set and dilated sparsity patterns when modeling such data.

\label{sec:conclusions}



\bibliographystyle{splncs} 
\bibliography{bibliography}

\begin{thebibliography}{10}

\bibitem{Ledig2017SRGAN}
Ledig, C., Theis, L., Huszar, F., Caballero, J., Cunningham, A., Acosta, A.,
  Aitken, A., Tejani, A., Totz, J., Wang, Z., Shi, W.:
\newblock Photo-realistic single image super-resolution using a generative
  adversarial network.
\newblock In: CVPR. (July 2017)

\bibitem{Yan2016VAEattr2img}
Yan, X., Yang, J., Sohn, K., Lee, H.:
\newblock Attribute2image: Conditional image generation from visual attributes.
\newblock In: ECCV. (2016)

\bibitem{Shu2017NeuralEdit}
Shu, Z., Yumer, E., Hadap, S., Sunkavalli, K., Shechtman, E., Samaras, D.:
\newblock Neural face editing with intrinsic image disentangling.
\newblock In: CVPR. (July 2017)

\bibitem{Vincent2008DAE}
Vincent, P., Larochelle, H., Bengio, Y., Manzagol, P.A.:
\newblock Extracting and composing robust features with denoising autoencoders.
\newblock In: ICML. (2008)  1096--1103

\bibitem{Kingma2014VAE}
Kingma, D.P., Welling, M.:
\newblock Auto-encoding variational bayes.
\newblock ICLR (2014)

\bibitem{Rezende2014VAE_DGLM}
Rezende, D.J., Mohamed, S., Wierstra, D.:
\newblock Stochastic backpropagation and approximate inference in deep
  generative models.
\newblock ICML (2014)

\bibitem{Dorta2018CVPR}
Dorta, G., Vicente, S., Agapito, L., Campbell, N.D.F., Simpson, I.:
\newblock Structured uncertainty prediction networks.
\newblock In: CVPR. (2018)

\bibitem{Yu2015Lsun}
Yu, F., Zhang, Y., Song, S., Seff, A., Xiao, J.:
\newblock Lsun: Construction of a large-scale image dataset using deep learning
  with humans in the loop.
\newblock arXiv preprint arXiv:1506.03365 (2015)

\bibitem{Rezende2015NF}
Rezende, D.J., Mohamed, S.:
\newblock Variational inference with normalizing flows.
\newblock ICML (2015)

\bibitem{Kingma2016IAF}
Kingma, D.P., Salimans, T., Welling, M.:
\newblock Improving variational inference with inverse autoregressive flow.
\newblock NIPS (2016)

\bibitem{Goodfellow2014GAN}
Goodfellow, I., Pouget-Abadie, J., Mirza, M., Xu, B., Warde-Farley, D., Ozair,
  S., Courville, A., Bengio, Y.:
\newblock Generative adversarial nets.
\newblock In: NIPS. (2014)  2672--2680

\bibitem{Makhzani2016AdversarialAE}
Makhzani, A., Shlens, J., Jaitly, N., Goodfellow, I.:
\newblock Adversarial autoencoders.
\newblock In: ICLR. (2016)

\bibitem{Mescheder2017AVB}
Mescheder, L., Nowozin, S., Geiger, A.:
\newblock Adversarial variational bayes: Unifying variational autoencoders and
  generative adversarial networks.
\newblock In: ICML. (2017)

\bibitem{Burda2016IWAE}
Burda, Y., Grosse, R., Salakhutdinov, R.:
\newblock Importance weighted autoencoders.
\newblock ICLR (2016)

\bibitem{Pu2017Stein}
Pu, Y., Gan, Z., Henao, R., Li, C., Han, S., Carin, L.:
\newblock Vae learning via stein variational gradient descent.
\newblock NIPS (2017)

\bibitem{Larsen2015VAEGAN}
Larsen, A.B.L., S{\o}nderby, S.K., Winther, O.:
\newblock Autoencoding beyond pixels using a learned similarity metric.
\newblock In: ICML. Volume~48., JMLR (2016)  1558--1566

\bibitem{Radford2016DCGAN}
Radford, A., Metz, L., Chintala, S.:
\newblock Unsupervised representation learning with deep convolutional
  generative adversarial networks.
\newblock ICLR (2016)

\bibitem{Berthelot2017BEGAN}
Berthelot, D., Schumm, T., Metz, L.:
\newblock Began: Boundary equilibrium generative adversarial networks.
\newblock CoRR (2017)

\bibitem{Arjovsky2017WGAN}
Arjovsky, M., Chintala, S., Bottou, L.:
\newblock Wasserstein gan.
\newblock CoRR (2017)

\bibitem{Oord2016PixelRNN}
Oord, A., Kalchbrenner, N., Kavukcuoglu, K.:
\newblock Pixel recurrent neural networks.
\newblock ICML (2016)

\bibitem{Chen2017LossyVAE}
{Chen}, X., {Kingma}, D.P., {Salimans}, T., {Duan}, Y., {Dhariwal}, P.,
  {Schulman}, J., {Sutskever}, I., {Abbeel}, P.:
\newblock {Variational Lossy Autoencoder}.
\newblock In: ICLR. (2017)

\bibitem{Gulrajani2017PixelVAE}
Gulrajani, I., Kumar, K., Ahmed, F., Taiga, A.A., Visin, F., Vazquez, D.,
  Courville, A.:
\newblock {PixelVAE: A Latent Variable Model for Natural Images}.
\newblock In: ICLR. (2017)

\bibitem{Makhzani2017PixelGAN_AE}
Makhzani, A., Frey, B.J.:
\newblock Pixelgan autoencoders.
\newblock In: NIPS.
\newblock (2017)  1975--1985

\bibitem{nikias1988}
Nikias, C.L., Pan, R.:
\newblock Time delay estimation in unknown gaussian spatially correlated noise.
\newblock IEEE Transactions on Acoustics, Speech, and Signal Processing
  \textbf{36}(11) (1988)  1706--1714

\bibitem{WOOLRICH2001}
Woolrich, M.W., Ripley, B.D., Brady, M., Smith, S.M.:
\newblock Temporal autocorrelation in univariate linear modeling of fmri data.
\newblock NeuroImage \textbf{14}(6) (2001)  1370 -- 1386

\bibitem{rasmussen2006}
Rasmussen, C.E., Williams, C.K.:
\newblock Gaussian processes for machine learning. Volume~1.
\newblock MIT press Cambridge (2006)

\bibitem{kendall2017}
Kendall, A., Gal, Y.:
\newblock What uncertainties do we need in bayesian deep learning for computer
  vision?
\newblock In: NIPS. (2017)

\bibitem{Liu2015CelebADataset}
Liu, Z., Luo, P., Wang, X., Tang, X.:
\newblock Deep learning face attributes in the wild.
\newblock In: ICCV. (December 2015)

\bibitem{Tensorflow}
Abadi, M., Agarwal, A., Barham, P., Brevdo, E., Chen, Z., Citro, C., Corrado,
  G.S., Davis, A., Dean, J., Devin, M., et~al.:
\newblock {TensorFlow}: Large-scale machine learning on heterogeneous systems
  (2015) Software available from tensorflow.org.

\bibitem{Kingma2015Adam}
Kingma, D., Ba, J.:
\newblock Adam: A method for stochastic optimization.
\newblock ICLR (2015)

\bibitem{Higgins2017BVAE}
Higgins, I., Matthey, L., Pal, A., Burgess, C., Glorot, X., Botvinick, M.,
  Mohamed, S., Lerchner, A.:
\newblock beta-vae: Learning basic visual concepts with a constrained
  variational framework.
\newblock In: ICLR. (2017)

\end{thebibliography}

\end{document}